\documentclass{article}

\usepackage{PRIMEarxiv}

\usepackage[utf8]{inputenc} 
\usepackage[T1]{fontenc}    
\usepackage{hyperref}       
\usepackage{url}            
\usepackage{booktabs}       
\usepackage{amsfonts}       
\usepackage{nicefrac}       
\usepackage{microtype}      
\usepackage{lipsum}
\usepackage{fancyhdr}  
\usepackage{algorithm,algpseudocode}
\usepackage{caption}
\usepackage{graphicx}   
\usepackage{multirow}
\usepackage{commath}

\graphicspath{{media/}}     

\begin{document}

\title{Minority Class Oriented Active Learning for Imbalanced Datasets}

\author{Umang Aggarwal$^{1,2}$, Adrian Popescu$^{1}$, Céline Hudelot$^{2}$\\
\small{(1) Université Paris-Saclay, CEA, Département Intelligence Ambiante et Systèmes Interactifs}\\
\small{(2) 
Université Paris-Saclay, CentraleSupélec, Mathématiques et Informatique pour la Complexité et les Systèmes} \\
\small{91191 Gif-sur-Yvette, France}\\
{\tt\small {umang.aggarwal,adrian.popescu}@cea.fr,celine.hudelot@centralesupelec.fr}
}

\maketitle

\begin{abstract}
Active learning aims to optimize the dataset annotation process when resources are constrained.
Most existing methods are designed for balanced datasets.
Their practical applicability is limited by the fact that a majority of real-life datasets are actually imbalanced.
Here, we introduce a new active learning method which is designed for imbalanced datasets.
It favors samples likely to be in minority classes so as to reduce the imbalance of the labeled subset and create a better representation for these classes. 
We also compare two training schemes for active learning: (1) the one commonly deployed in deep active learning using model fine tuning for each iteration and (2) a scheme which is inspired by transfer learning and exploits generic pre-trained models and train shallow classifiers for each iteration.
Evaluation is run with three imbalanced datasets.
Results show that the proposed active learning method outperforms competitive baselines.
Equally interesting, they also indicate that the transfer learning training scheme outperforms model fine tuning if features are transferable from the generic dataset to the unlabeled one. 
This last result is surprising and should encourage the community to explore the design of deep active learning methods.
\end{abstract}

\section{Introduction}

Efficient deep learning relies on the availability of large annotated datasets which are exploited to optimize the large number of parameters of deep neural networks.
The success of fully supervised strategies is conditioned by sizes of the networks and of the available annotated datasets. 
Large amounts of unlabeled data are available for many tasks but their annotation can be costly.  
Partial annotation of datasets can be an interesting solution for time-sensitive applications or when annotation resources are constrained.
In such cases, the effectiveness of the training process is linked to the quality of the sampling performed. 

Active Learning (AL) \cite{Settles10activelearning} methods are designed to optimize the accuracy of model learned over annotated subset of unlabeled dataset. 
Given a total annotation budget, AL is usually deployed in an iterative fashion.
A fixed number of samples are selected per iteration and annotated in order to retrain the model which become gradually stronger.
Samples are selected according to an acquisition function (AF) whose objective is to maximize informativeness~\cite{Shannon1948,DBLP:conf/aaai/CulottaM05,DBLP:conf/icdm/SchefferDW01,DBLP:conf/cvpr/BeluchGNK18},  representativeness~\cite{DBLP:conf/iclr/SenerS18, DBLP:conf/icmla/LiGC12,DBLP:conf/icml/DasguptaH08} or a combination of these two criterias~\cite{DBLP:journals/pami/ChakrabortyBSPY15}.
Measures used to maximise informativeness prioritize data for which the model is least certain. 
Measures based on representativeness focus on selecting a diverse set of samples so as to provide a strong representation of the unlabeled dataset. 

Machine learning algorithms are often designed and evaluated under the assumption that datasets are balanced or nearly so.
However, in practice, data are often unevenly distributed across classes and imbalance needs to be dealt with~\cite{DBLP:journals/nn/BudaMM18}.
Moreover, the classes of interest are often the ones in minority in a range of applications such as the medical domain~\cite{rao2006data} and  big data processing~\cite{8614150}.
Classifiers trained on imbalanced datasets are likely to be overwhelmed by samples coming from majority classes.
The more general focus on learning with balanced datasets is also observed in AL, with a limited number of existing works addressing imbalance. 
It is argued in~\cite{ertekin2007active} that AL can be adapted to class imbalance by focusing on samples near the class boundaries where the imbalance is observed to be lower compared to the overall distribution. 
This observation and a margin exhaustion criterion are employed to limit the selection of samples from majority classes. 
However, the authors of~\cite{attenberg2013class} show that a high degree of imbalance has an adverse effect on the selection process. 
The resulting model will be biased towards selecting samples from majority classes. 
Also, the prioritization of samples which are close to the hyperplane can have negative effects, as it fails to create a good representation of the minority class. 
Particularly, the problem can be aggravated when the minority class contains several concepts or is not easily separable from a majority class~\cite{attenberg2013class,he2008nearest}. 
Consequently, a diverse representation of minority classes should be targeted.


We introduce a new method which tackles imbalanced AL by focusing on samples which are classified as minority by the model learned in the previous AL iteration. 
While simple, this approach has two advantages.
First, if the samples are correctly classified as minority classes, the selection of these samples mitigates imbalance in the labeled subset and results in a better representation of the minority classes. 
Second, if the samples are mis-classified as minority, we hypothesize that these samples have high informative value. 
Indeed, in this case, the model learns from samples which it previously mis-classified, thereby adding important missing information.
It also prunes the decision boundaries around the minority classes as the samples mis-classified as minority class are likely to be somewhat in the vicinity of the minority class. their mis-classified samples are located in their vicinity in the representation space.
We use three selection strategies to sample from minority class based on uncertainty, certainty or diversity, respectively.
These intra-class selection strategies mirror the usual AL selection strategies which are applied at the dataset level.
The minority status of a class is dynamically assigned after each iteration by updating statistics about the class distribution. 
The number of samples for a minority class is based on the degree of imbalance in the class.

Note that minority class predictions might not be numerous enough to cover the entire AL iteration. 
If so, the three strategies described above become equivalent since all minority samples will be selected.
Then, the remaining budget of the iteration will be selected using a classical acquisition function, such as random or margin sampling. 

A majority of deep AL works exploit iterative fine tuning~\cite{DBLP:conf/iclr/SenerS18,DBLP:conf/cvpr/ZhouSZGGL17} to progressively learn stronger deep models.
In~\cite{Aggarwal_2020_WACV}, we showed that the use of a pre-trained model and of SVM classifiers is preferable to fine tuning in the early stage of single-state AL if features are transferable toward the current task.
Beyond the proposal of a new AL method, we provide a comparison of the two learning strategies in iterative AL and propose a combination of them to maximize accuracy.

Evaluation is done with three imbalanced datasets designed for different visual tasks.
Results indicate that the proposed method outperforms four competitive baselines.
In addition to global results, we present an analysis of the method components so as to understand their individual roles.

\section{Related Works}

\subsection{Active learning}
A large number of methods have been proposed in AL literature to select the most pertinent samples for manual labeling~\cite{Settles10activelearning}. 
Their objective is to maximize either informativeness or representativeness of the manually labeled subset.
Informativeness is targeted by favoring the selection of samples about which the classification model is most uncertain.
The rationale for this approach is to improve the learned model by focusing it on difficult regions of the representation space which are close to classifier boundaries.
Least confidence first~\cite{DBLP:conf/aaai/CulottaM05}, margin sampling~\cite{DBLP:conf/icdm/SchefferDW01} or entropy~\cite{Shannon1948} are the most common methods  which measure uncertainty.
The main limitation of informativeness-based approaches is that they fail to capture the overall sample distribution and hence provide weak representation of each class. 
This problem is more acute at the start of the iterative learning cycle when the uncertainty measures are less reliable as the class representations are still incomplete~\cite{Settles10activelearning}. 
This can be particularly problematic for minority classes as their representations would be weaker compared to those of majority classes. Then, samples from minority classes could be easily mis-classified with confidence in any of the majority classes.

Representativeness-based approaches consist in increasing the diversity of selected samples.
They aim to select the subset such that it best represents the true distribution.
Several works focus on selecting samples based on K-means and hierarchical clustering based approaches~\cite{DBLP:conf/icmla/LiGC12},~\cite{DBLP:conf/icml/DasguptaH08}.
Their objective is to distribute the selected samples across different regions in the feature space.
We incorporate this idea in the context of class imbalance, to focus on minority classes by selecting samples from regions associated with minority classes. 
However, sample selection is done in the class prediction space instead of the feature space used for clustering.
Coreset~\cite{DBLP:conf/iclr/SenerS18} is a recent approach that takes a min-max view to select the samples whose distance to the closest labeled sample is the maximum.  
In our work, we use Coreset to select a diverse set of samples among those associated with minority classes.

AL has recently received increasing attention in the context of deep learning. 
One line of works tries to improve the uncertainty measures.
Several runs of the models with random dropout parameters are exploited in Monte Carlo dropout~\cite{DBLP:conf/icml/GalIG17} to generate more stable uncertainty measures. 
An ensemble strategy is deployed in~\cite{DBLP:conf/cvpr/BeluchGNK18} by exploiting the snapshot of the model at different iteration points in the training process to generate uncertainty measures. 
These approaches are complementary to our work.
Note that while effective, their application makes the AL process more complex since multiple versions of deep models are needed.  

A cold start problem appears when one tries to perform deep active learning for an unlabeled dataset.
The random selection of an initial sample set for labeling is usually proposed in AL~\cite{DBLP:conf/cvpr/BeluchGNK18,DBLP:conf/iclr/SenerS18,DBLP:conf/icml/GalIG17}.
The effectiveness of this approach depends on the size of the initial set. this is particularly the case for large deep learning models, that might provide unstable uncertainty measures at small budgets. This problem was tackled in~\cite{Coleman2020Selection}, where a small proxy model is used to perform sample selection before training a large model with sufficient data.  

An alternative is to use a transfer learning strategy~\cite{DBLP:conf/cvpr/RazavianASC14} to avoid cold start~\cite{DBLP:conf/cvpr/ZhouSZGGL17,Aggarwal_2020_WACV}. 
A model pre-trained on a generic dataset is used as feature extractor and shallow classifiers are trained with these features for the current dataset. 
We follow a similar approach by training shallow classifiers over pre-trained features and compare their use to the more usual iterative fine tuning.

\subsection{Class imbalance learning}
Imbalanced learning is a well-studied problem in literature. 
Various groups of methods were devised to tackle it.
They focus either on data sampling or classifier optimization \cite{DBLP:journals/tkde/HeG09}.
Imbalance can be reduced by either random oversampling of under-represented minority classes or with random undersampling of over-represented majority classes \cite{DBLP:journals/eswa/GuoLSMYB17}.
Both methods have their limitations as oversampling can lead to overfitting, whereas undersampling can lead to incomplete representation of majority classes. 
Sampling methods are shown to have limited or even detrimental improvement in deep learning context~\cite{DBLP:journals/nn/BudaMM18} and are thus not in focus here. 

Another line of work deals with imbalance at the classifier level~\cite{DBLP:journals/tkde/HeG09}. 
Class weights can be incorporated in the loss function during training.
Alternately, a post processing such as thresholding is applied to rectify class predictions during test based on their prior probabilities in the training set. 
While simple, thresholding was shown to outperform a large array of data sampling and classifier level methods for object recognition using deep learning models in \cite{DBLP:journals/nn/BudaMM18}. 
We incorporate thresholding in the fine tuning based training scheme.
Cost-sensitive weighted SVM acts as an effective method to find the optimal hyperplanes for imbalanced classes in SVMs~\cite{fernandez2018learning}. 
The hinge loss of samples is modified based on the number of samples per class to give more importance to minority classes. 
We use this approach when deploying AL with SVMs learned over a pretrained model. 

\subsection{Active learning in class imbalance learning}
The authors of~\cite{ertekin2007active} concluded that samples close to the decision boundaries are likely to have less imbalance than the overall distribution and this observation is used to drive sample balancing.
Uncertain samples are selected in~\cite{ertekin2007active,zhu2007active} along with a margin exhaustion criteria to limit the selection for majority classes. 
However, it fails to provide a strong representation of minority classes as selection process is focused only on samples within the margins of hyperplanes~\cite{attenberg2013class}.
Note that cost-sensitive SVM (CS-SVM) was exploited as an effective way to handle a skewed data distribution during active learning~\cite{ertekin2007active,zhu2007active}. 
Other works try to explicitly favor the selection of minority class samples to reduce imbalance.  
The mis-classification component is used to penalize selection of majority classes with boosted SVMs in~\cite{zikeba2015boosted}. 
In~\cite{he2008nearest}, the authors propose to prioritize samples which are nearest neighbours of minority classes. 
Note that most of these works are designed for binary classification and are not easily transferable to the multi-class classification problem tackled here. 

In~\cite{Aggarwal_2020_WACV}, we tackled the cold start in AL for imbalanced datasets using a single-stage scenario.
The selection of an initial diversified and balanced sample set was done using a pretrained model and a representativeness-oriented acquisition function.
Put simply, a sample was labeled if it was close to a minority class and far from any majority class.
The main differences between the approach proposed here and the one in~\cite{Aggarwal_2020_WACV} are: (1) the proposal of a different acquisition function, (2) the use of a more generic iterative AL setting and (3) the adaptation of shallow classifiers to an imbalanced context. 

\section{Problem Formalization}
\label{sec:formalization}
We consider an unlabelled dataset $\mathbb{D}^{U}$ with samples $x_i \in \mathcal{X}$ for $i = [1..u] $, i.i.d realizations of random variables $\mathcal{X}$ drawn from the distribution $\mathbb{P}$ where $\mathcal{X}$ is the instance space. 
In an iterative AL setting, a total budget $b$, with $b < u$, is allocated for manual labeling in $t$ iterations.
The process starts by randomly selecting a small subset of $\mathbb{D}^{U}$ for annotation to create the initial labeled dataset $\mathbb{D}_{0}^{L}$ with $x_j,y_j \in \mathcal{X} \times \mathcal{Y}$  for $j = [1..\frac{b}{t}]$.
Here 
$\mathcal{Y} = \{y_1,...,y_n\}$ is the set of $n$ class labels.
$\mathbb{D}_{0}^{L}$ is used to train an initial model $\mathcal{M}_{0}$ which provides the probability estimates $\mathcal{P}_{0}$  or the feature embeddings $\mathcal{F}_{0}$ required by an acquisition function $AF$ to estimate the importance of samples for the next iteration.
Afterwards, at iteration step $k$, for $k = [1..t-1]$, a batch of samples of size $\frac{b}{t}$ is selected for labeling from $\mathbb{D}_{k}^{U} = \mathbb{D}^{U} \setminus \mathbb{D}_{k-1}^{L}$, and added to $\mathbb{D}_{k-1}^{L}$ to update the labeled subset $\mathbb{D}_{k}^{L}$. 
$\mathbb{D}_{k}^{L}$ is then used to learn the model $\mathcal{M}_{k}$.

We test two training methods which are based either on deep model update or on a pretrained model followed by a shallow learning of classifiers during AL. 
Both schemes exploit a pretrained model $\mathcal{M}_{S}$ learned over a generic dataset.
The first method is inspired from the usual deep AL process and is based on fine tuning~\cite{DBLP:conf/cvpr/BeluchGNK18,DBLP:conf/iclr/SenerS18,DBLP:conf/icml/GalIG17}.
Deep models are fine tuned for each AL iteration using $\mathcal{M}_{S}$ and $\mathbb{D}_{k}^{L}$.
The second method assumes that, although $\mathcal{M}_{S}$ is trained on a separate dataset, the knowledge it encapsulates can be transferred to $\mathbb{D}^{U}$.
Feature embeddings $\mathcal{F}_{S}$ from $\mathcal{M}_{S}$ are used to train shallow classifiers during AL iterations.
We propose a switch criterion to decide at which iteration fine tuning should be used instead of transfer learning AL scheme.
This criterion is implemented using cross-validation at the end of each iteration to assess the performance of each of the two training schemes. 

\section{Active learning acquisition Process}

\subsection{Baselines}
We compare our method to four competitive baselines. The first three are classical methods representative for random, informativeness-based, and diversity-based selection respectively.
The fourth is a method which we recently introduced to counter cold start in imbalanced AL~\cite{Aggarwal_2020_WACV}.

\subsubsection{Random sampling}
The simplest baseline consists in a random selection of samples for annotation.
Note that this sampling roughly reproduces dataset imbalance. 
It is a good reference to measure how effective imbalance countering is for each of the other methods tested. This baseline is noted $random$ below.

\subsubsection{Margin sampling}
Uncertainty based methods aim to maximize informativeness and thus focus on samples that the model is having most difficulty in classifying. 
They are deployed using the outputs of the classification layer~\cite{Settles10activelearning}.
Margin based AL is shown to be an effective uncertainty measure for class imbalance learning problems~\cite{ertekin2007active}. It exploits the region between two different classes where imbalance is also likely to be limited.
It is defined as:
\begin{equation}
 marg(x) =   (P(y = c^1|x) - P(y = c^2|x))   
 \label{eq:ms}
\end{equation}
where $c^1,c^2$ are the top-2 predicted classes for test sample $x$.

Samples are prioritized if the difference between the top two predictions given by Equation~\ref{eq:ms} is low.
The baseline obtained is noted $margin$.

\subsubsection{Coreset}
Diversity based methods aim to select a subset of $\mathbb{D}^U$ so as to ensure an optimal coverage of the representation space.
Coreset~\cite{DBLP:conf/iclr/SenerS18} was recently introduced as a way to solve the greedy k-center problem. 
It is defined as:
\begin{equation}
  core(\mathbb{D}_{k}^{U},\mathbb{D}_{k}^{L}) = 
    \max_{\forall x_u \in \mathbb{D}_{k}^{U}} \min_{x_l \in \mathbb{D}_{k}^{L} } d(F(x_u),F(x_l))  
\label{eq:core}
\end{equation}
where $core(\mathbb{D}_{k}^{U},\mathbb{D}_{k}^{L})$  returns a samples from unlabeled dataset $\mathbb{D}_{k}^{U}$ using the labeled dataset  $\mathbb{D}_{k}^{L}$,  $d(F(x_u),F(x_l))$ is the distance between a labeled point $x_l$ from $\mathbb{D}_{k}^{L}$ and an unlabeled point $x_u$ from $\mathbb{D}_{k}^{U}$. $F$ is the feature extractor. For SVM training, a classifier is learnt with $F$ = $F_S$ given by the pretrained model $\mathcal{M}_S$. 
For the CNN fine tuning, the feature extractor is given by $\mathcal{M}_{k-1}$, the model available from the latest iteration. 
In Equation~\ref{eq:core}, the unlabeled samples which sit at minimum distance from the k-centers already labeled are first preselected. 
Then, the sample among the minima which is at a maximum distance from any of the k-centers is kept. The k-centers are updated after every selected.
The baseline AF obtained by applying Equation~\ref{eq:core} is noted $coreset$.

\subsubsection{Certainty diversified sampling}
We introduced this function, abbreviated $cds-bal$ below, in~\cite{Aggarwal_2020_WACV} to deal with the cold start problem in single-stage imbalanced AL.
It exploits a pretrained model $\mathcal{M}_S$ to provide the features for its diversification and balancing objectives.
Note that $cds-bal$ is deployed in the feature space so as to select samples which minimize their distance to the centroid of a minority class and to maximize the distance to the centroid of the closest majority class.
$cds-bal$ is adapted here for usage in an iterative AL setting by selecting the initial subset using random sampling as in other AL acquisition function.

\subsection{Method}
\textbf{Minority class oriented sample acquisition.}
\label{subsec:method}
Minority classes have weaker representations in the models trained from imbalanced datasets.
They need to be prioritized either during training or test to reduce the effect of imbalance~\cite{DBLP:journals/nn/BudaMM18,DBLP:journals/tkde/HeG09}.
We translate this observation to an iterative AL scenario to propose a simple and efficient method which improves sampling from imbalanced datasets.
Minority classes are identified by computing statistics of the class distribution of labeled data points up to the last iteration. 
This distribution also provides the estimated number of samples needed in the current iteration to remove imbalance for each minority class.
The initial set of candidates for a minority class is made of samples predicted as belonging to it by $\mathcal{M}_{k-1}$, the latest model learned.
Selection of candidate samples can be done to boost certainty, uncertainty or diversity and leads to the different versions of the proposed AF discussed below.

At the start of the $k^{th}$ iteration, the number of labeled samples is $s_{k} = k \frac{b}{t}$ based on iterations $0$ to $k-1$.
The objective is to add $\frac{b}{t}$ new samples with priority given to minority classes.
We note $s_{k}^c$, the number of labeled samples for class $c$ and compute the average number of samples per class $\mu_k = \frac{s_k}{n}$.
A class is then considered as minority if $s_k^c < \mu_k$.
The maximum number of samples which is allowed for class $c$ during the $k^{th}$ iteration is defined as:

\begin{equation}
   m_k^c= 
\begin{cases}
    \mu_k - s_k^c,& \text{if } s_k^c < \mu_k\\
    0,              & \text{otherwise}
\end{cases}
\label{eq:selection}
\end{equation}

Equation~\ref{eq:selection} favors minority classes since they are the only ones which have candidate samples allocated.
The set of unlabeled samples associated to class $c$ is given by:
\begin{equation}
    \mathbb{D}^{U(k)}_c = \{\forall x \in \mathbb{D}_{k}^{U}, if \: P(c^1 = c|x) \}
\label{eq:samples}
\end{equation}
, where $c^1$ is the predicted label for the sample $x$ and.

If $ \vert \mathbb{D}^{U(k)}_c \vert > m_k^c$, a selection is needed among the set of samples given by Equation~\ref{eq:samples}.
We propose three ways to select samples which are inspired from classical AL objectives.

\subsubsection{Certainty-oriented Minority Class Sampling} favors the most certain data points from $\mathbb{D}^{U(k)}_c$ using:
\begin{equation}
     CMCS = arginvsort_{\forall x \in \mathbb{D}^{U(k)}_c} marg(x) 
\label{eq:CMCS}
\end{equation}
where $marg(x)$ is the margin measure from Equation~\ref{eq:ms} and $arginvsort$ sorts the samples in decreasing order.  
Note that Equation~\ref{eq:CMCS} performs a margin sampling at class level instead of dataset level. 
It thus allows a selection of certain samples but oriented towards minority classes due to the application of Equation~\ref{eq:selection} for sample allocation to classes.

\subsubsection{Uncertainty-oriented Minority Class Sampling} favors the most uncertain samples from $\mathbb{D}^{U(k)}_c$ using:
\begin{equation}
     UMCS= argsort_{\forall x \in \mathbb{D}^{U(k)}_c} marg(x)
\label{eq:UMCS}
\end{equation}
where $marg(x)$ the margin from Equation~\ref{eq:ms} and $argsort$ sorts the samples in increasing order.
Equation~\ref{eq:UMCS} favors data points which are predicted under $c$ but are close to other classes. 
Its objective is inverse compared to that of $CMCS$.

\subsubsection{Diversity-oriented Minority Class Sampling} It aims to select a diversified samples subset for $c$. 
Such a subset can be obtained, for instance, by applying the Coreset method~\cite{DBLP:conf/iclr/SenerS18} from Equation~\ref{eq:core} to $\mathbb{D}^{U(k)}_{c}$:
\begin{equation}
  DMCS   = core( \mathbb{D}^{U(k)}_c, \mathbb{D}^{L(k)}_c)     
\label{eq:DMCS}
\end{equation}
where $core$ is applied iterative-ly to select $m^c_k$ samples from $\mathbb{D}^{U(k)}_c$ using the samples already selected $\mathbb{D}^{L(k)}_c$ for class $c$, while updating the datasets with every selection.

\textbf{Auxiliary acquisition functions}
The proposed sampling process is focused on minority classes. 
It is possible that if the imbalance is limited the number of samples allocated to minority classes is less than the budget. Further, there is no guarantee that there are enough samples predicted under minority classes to treat the imbalance. 
Minority predicted samples are most likely to be insufficient at either at the very beginning or towards the end of the AL process.
In the beginning, minority class representations are weak and their samples are likely to be mis-classified in majority classes.
Towards the end, there will be simply too few samples left for labeling in minority classes.  
In such cases if the budget of the $k^{th}$ iteration is not filled entirely, remaining samples can be selected according to any AF.
Tests are run using random and margin sampling for these remaining samples.
The final forms proposed acquisition functions are noted $DMCS-rand$ and $DMCS-marg$.

\section{Experiments}
We first describe the experimental setup and the datasets used in evaluation.
Then, we discuss the obtained results globally and also present an analysis of the main components of the proposed approach.
\subsection{Setup}
We test the proposed approach using an usual iterative AL setting~\cite{DBLP:conf/cvpr/BeluchGNK18,DBLP:conf/iclr/SenerS18,DBLP:conf/icml/GalIG17}.
We set the AL budget to $b=8000$ and the number of iterations to $t=16$, including the initial one.
The number of samples selected in each iteration is $500$.

We use a ResNet-18 architecture~\cite{DBLP:conf/cvpr/HeZRS16} for all experiments.  
The ResNet-18 model is trained over the ILSVRC dataset~\cite{DBLP:journals/ijcv/RussakovskyDSKS15} is used $\mathcal{M}_S$.

As discussed in Section~\ref{sec:formalization}, AL performance is tested with two training schemes. 
The first scheme is based on fine tuning as proposed in previous deep AL works~\cite{DBLP:conf/cvpr/BeluchGNK18,DBLP:conf/iclr/SenerS18,DBLP:conf/icml/GalIG17}. 
We employ thresholding based on prior class probabilities to reduce the effects of imbalance~\cite{DBLP:journals/nn/BudaMM18}.  
This scheme is noted $FT-th$ and is used by default in experiments. $FT-th$ models are trained for 60 epochs with an initial learning rate of 0.01 and a batch size of 32. The Stochastic gradient descent was used with the cross-entropy loss. A learning rate decay of 0.1 was done if the loss plateaus for 10 epochs.
The second scheme is inspired from transfer learning and exploits a model pretrained on ILSVRC.
It is less frequent in deep active learning but proved useful to tackle cold start problem~\cite{Aggarwal_2020_WACV}.
SVMs are trained after each iteration using the features provided by $\mathcal{M}_S$, the pretrained model.
Following~\cite{ertekin2007active,zhu2007active}, cost-sensitive SVMs are used to reduce the negative effect of imbalance.
This scheme is noted $CS-SVM$ and is used by default in experiments.
Results obtained with fine tuning without thresholding (noted $FT$) and with classical SVMs (noted $SVM$) are also reported to highlight the usefulness of adapting training schemes to an imbalanced learning context.
The two training schemes are run in parallel at the start of the AL process in order to exploit the one which is more accurate.
Transfer learning with SVMs is more likely to be useful at the beginning, until the $\mathbb{D}^L$ is sufficiently large for efficient training of deep models.
The switch between the two schemes will occur faster if the content of the unlabeled dataset is visually unrelated to the one in the generic model used for the pretrained model. 
Cross validation using 80:20 split is performed for each of the two schemes after each iteration.
The average accuracy of each scheme is computed to decide which of them should be used starting from the following iteration.
The methods which are non-deterministic in the iterative setting, namely random sampling and the proposed method with auxiliary random sampling are repeated with 5 different seeds. 

It is common practice in imbalanced learning~\cite{DBLP:journals/nn/BudaMM18,DBLP:journals/tkde/HeG09} to evaluate performance over balanced test datasets.
This choice is also made here to give equal importance to each class irrespective of the class distribution in the training dataset.

\subsection{Datasets}
The proposed method and the baselines are evaluated on three imbalanced datasets designed for different visual tasks.
Following~\cite{Aggarwal_2020_WACV}, we induce imbalance in the publicly available CIFAR-100~\cite{Krizhevsky09learningmultiple} (object recognition) FOOD-101~\cite{bossard14} (fine-grained food recognition), MIT-67~\cite{DBLP:conf/cvpr/QuattoniT09} (indoor scene recognition).
An imbalance induction procedure was applied to all datasets using a target imbalance ratio to guide the pruning process. 
The imbalance ratio is defined as $ir = \frac{\sigma}{\mu}$, with $\sigma$ the standard deviation and $\mu$ the mean of images per class in the dataset.
The main statistics of the obtained datasets are provided in Table~\ref{tab:dataset}. 
Imbalance is similar across datasets to facilitate comparability of results. 

\begin{table}[]
    \begin{center}
    \resizebox{0.45\textwidth}{!}
    {
    \begin{tabular}{|c|c|c|c|c|c|}
        \hline
         Dataset & Class & Images & Mean($\mu$) & Std($\sigma$) & $ir$   \\ \hline
         FOOD-101 &  101  & 22956  & 227.28 & 180.31   & 0.793 \\ \hline
         CIFAR-100 &  100  & 17168  & 171.68 & 126.98   & 0.740 \\ \hline 
         MIT-67   &   67  & 14281  & 213.15 & 168.16   & 0.789 \\ \hline
    \end{tabular}
    }
    \end{center}
    \caption{Dataset statistics. $ir$ is the imbalance ratio.}
    \label{tab:dataset}
\vspace{-8mm}
\end{table}

\begin{figure*}[ht]

\includegraphics[width=0.33\textwidth]{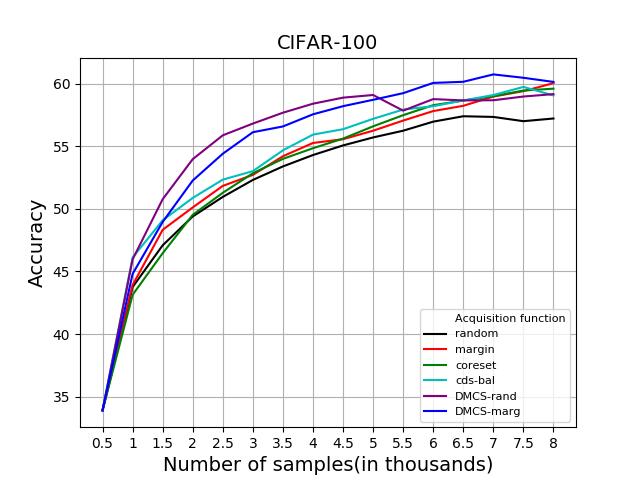}
\includegraphics[width=0.33\textwidth]{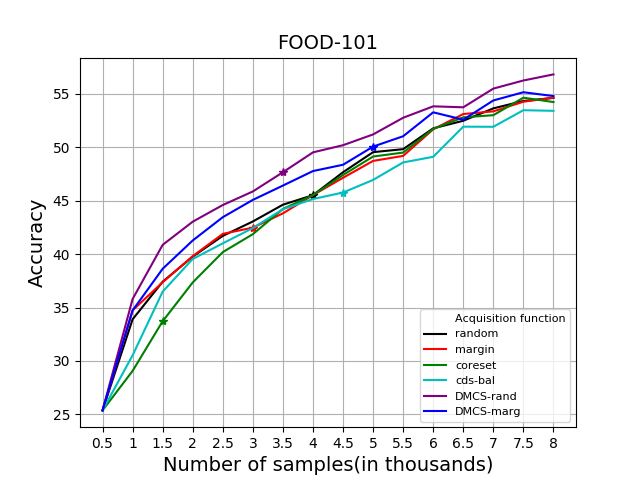}
\includegraphics[width=0.33\textwidth]{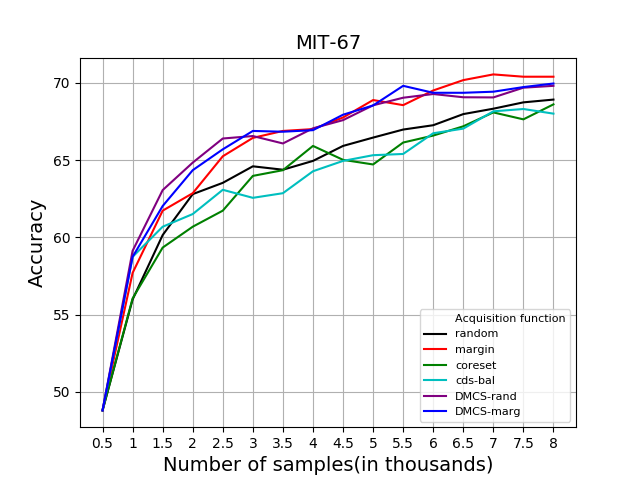}

\vspace{-1em}
    \caption{Iterative active learning performance for baselines and for the proposed method using cross-validation between $CS-SVM$ and $FT-th$ training schemes. Results with random ($rand$) and margin ($marg$) based sampling are shown for the remaining budget of each iteration when there are not enough samples associated to minority classes. "*" represents the switching point from $CS-SVM$ to $FT-th$ training scheme. The AL budget is $b=8000$ and the number of iterations $t=16$.  \textit{Best viewed in color}.}
    \label{fig:acc_full}
\vspace{-3mm}
\end{figure*}

\begin{figure*}[ht]

\includegraphics[width=0.33\textwidth]{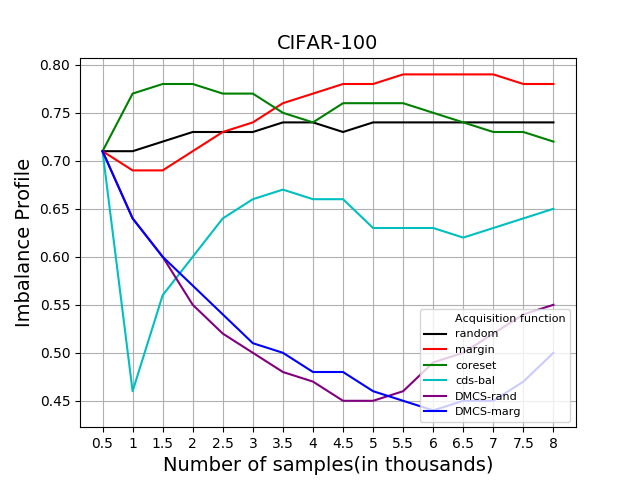}
\includegraphics[width=0.33\textwidth]{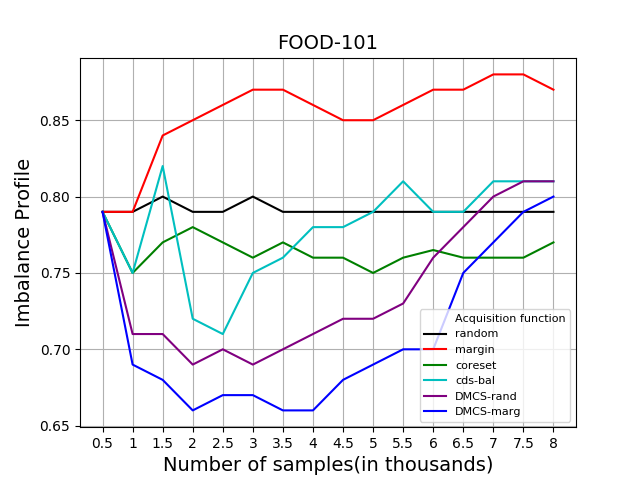}
\includegraphics[width=0.33\textwidth]{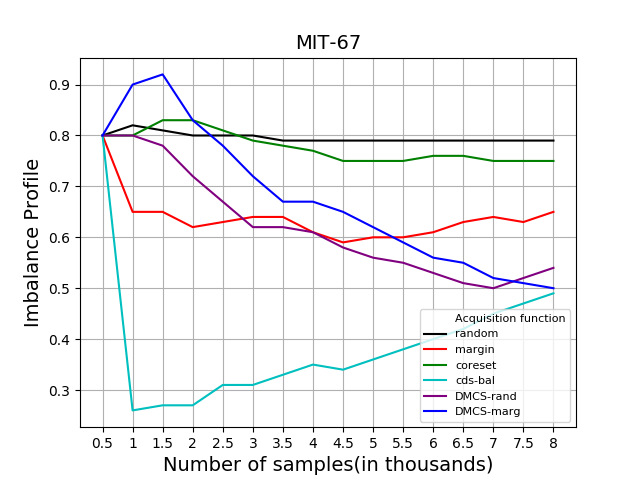}

\vspace{-1em}
    \caption{Imbalance profile of labeled datasets for different acquisition functions. $b=8000$, $t=16$.  \textit{Best viewed in color}.}
    \label{fig:imp_full}
\vspace{-4mm}
\end{figure*}  

\subsection{Global performance discussion}
The results obtained with the baseline methods and with $DMCS$, the diversified version of the proposed AF, are provided in Figure~\ref{fig:acc_full}.
A consistent performance gain is obtained with $DMCS-marg$ and $DMCS-rand$ compared to the baselines.
This indicates that the proposed method is appropriate for use in iterative AL for imbalanced datasets.
Accuracy improvements are obtained for all three datasets and are particularly interesting for $CIFAR-100$ and $FOOD-101$ datasets. 
The comparison of $DMCS-rand$ and $DMCS-marg$ is globally favorable to the first method and is discussed in more detail in Subsection~\ref{subsec:rand_margin}.
Mirroring global results, $DMCS-rand$ and $DMCS-marg$ have better performance for a large majority of individual iterations on $CIFAR-100$ and $FOOD-101$ from Figure~\ref{fig:acc_full}.
They are also better than $margin$ sampling for MIT-67 between 1000 and 3000 samples and results become more mixed afterwards as the uncertainty criteria becomes more reliable.
The better behavior of the proposed method is partly explained by its ability to select candidates for labeling whose distribution is globally more balanced, as illustrated in Figure~\ref{fig:imp_full}.
The imbalance profiles of the two $DMCS$ versions are clearly better than those of baselines for CIFAR-100 and FOOD-101. 
They are comparable to those of $margin$ for MIT-67.
It is noteworthy that the imbalance profile is not the only factor explaining AF performance.
This is clear from the analysis of $cds-bal$ imbalance profile, which is better than that of other methods but is not correlated with a performance gain.
The intrinsic quality of the selected samples is also important.
The reported results indicate that $DMCS$ is able to provide a more appropriate sampling than the other methods.

The performance of baselines methods is generally close to that of random sampling or even lower. The only exception is $margin$ for MIT-67, which is clearly better than $random$.
This result confirms previous reports~\cite{DBLP:conf/cvpr/BeluchGNK18,DBLP:conf/iclr/SenerS18} that random sampling is a competitive baseline in active learning. 
This is particularly the case for the imbalanced datasets tested here. 
$coreset$ is comparable to $random$ for $CIFAR-100$, but is inefficient for $FOOD-101$ and $MIT-67$. It also fails to provide any significant improvement in the imbalance profiles for the datasets. It is likely that, for imbalanced datasets, $coreset$ selects outliers that belong to majority classes.

$cds-bal$ is useful to tackle cold start in AL~\cite{Aggarwal_2020_WACV} but fails in selecting good quality samples in the iterative AL setting studied here. 

The results show that transfer from a generalist pretrained model is preferable at the beginning for all three datasets.
Surprisingly, this training scheme remains better than fine tuning for CIFAR-100 and MIT-67 throughout the entire AL process presented in Figure~\ref{fig:acc_full}.
As illustrated, the switch from SVM toward fine tuning occurs only for $FOOD-101$.
This is intuitive since this dataset was shown to be furthest away from $ILSVRC$ in~\cite{Aggarwal_2020_WACV}.
The performance of the two schemes is illustrated in detail in Figure~\ref{fig:ts} and further discussed in Subsection~\ref{subsec:classical_imbalanced}.
This finding is interesting insofar it is at odds with the usual assumption that fine tuning schemes should be used in iterative active learning~\cite{DBLP:conf/cvpr/BeluchGNK18,DBLP:conf/iclr/SenerS18,DBLP:conf/icml/GalIG17}.
It is also interesting because the transfer learning scheme is much faster since only shallow classifiers need to be trained for each iteration.

\subsection{Comparison of random and margin as auxiliary AFs}
\label{subsec:rand_margin}
In Figure~\ref{fig:acc_full}, $DMCS$ results are presented with $random$ and $margin$ as auxiliary AFs if there are not enough samples associated to minority classes.
It is somewhat surprising to note that $DMCS-rand$ provides slightly better overall accuracy compared to $DMCS-margin$. 
This happens even though $margin$ baseline is globally better than $random$ when used alone.
$DMCS-rand$ is better for all iterations for the $FOOD-101$ dataset although the imbalance profile in Figure~\ref{fig:imp_full} is better for $DMCS-marg$.
The difference between the two $DMCS$ variants is very small for $MIT-67$.
Their performance for $CIFAR-100$ is interesting insofar $DMCS-rand$ is more effective in up to 5000 samples and $DMCS-marg$ becomes better afterwards.
The change of performance is correlated with an inversion of imbalance profiles in Figure~\ref{fig:imp_full}.

We assume that some amount of randomness is effective in the beginning for driving the balancing procedure to focus sampling on minority classes.
There, random sampling provides a better overall representation than margin sampling. 
Later in the AL process, uncertainty estimates are more reliable and imbalance has been mitigated to the extent possible for the given dataset.
Then, it becomes preferable to select the remaining samples based on margin sampling.


\begin{figure*}[ht]

\includegraphics[width=0.33\textwidth]{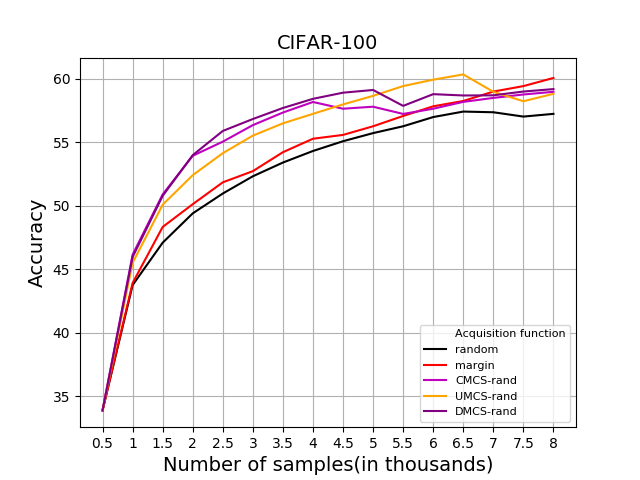}
\includegraphics[width=0.33\textwidth]{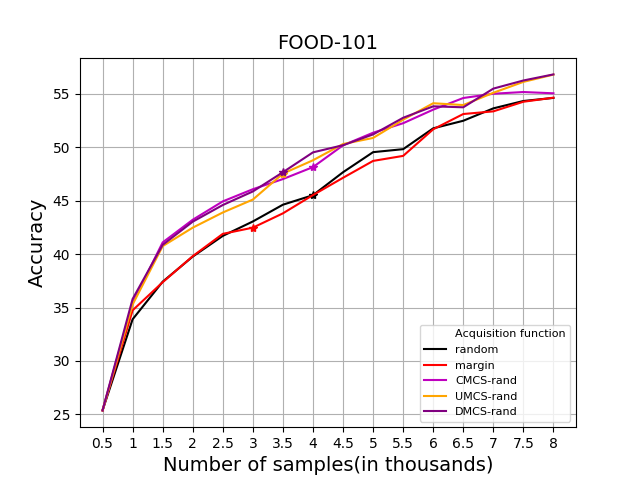}
\includegraphics[width=0.33\textwidth]{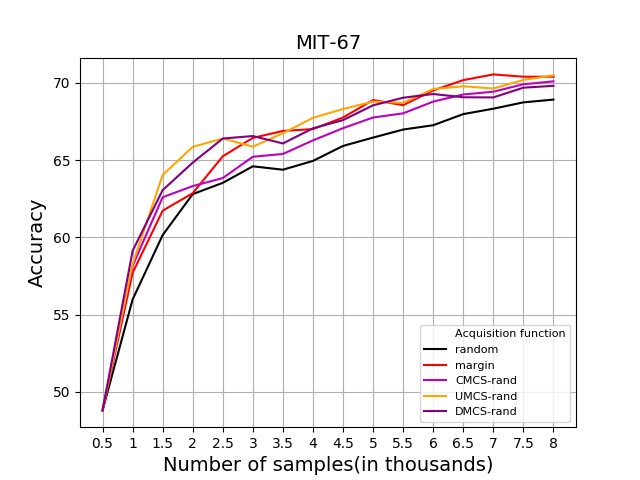}
    
\vspace{-1em}
    \caption{Iterative active learning performance of different versions of the proposed method. Random sampling is used as AF for remaining samples of each iteration if there are not enough samples associated to minority classes. $b=8000$, $t=16$.  \textit{Best viewed in color}.}
    \label{fig:acc_div}
    \vspace{-3mm}
\end{figure*}

\begin{figure*}[!htpb]
   
\includegraphics[width=0.33\textwidth]{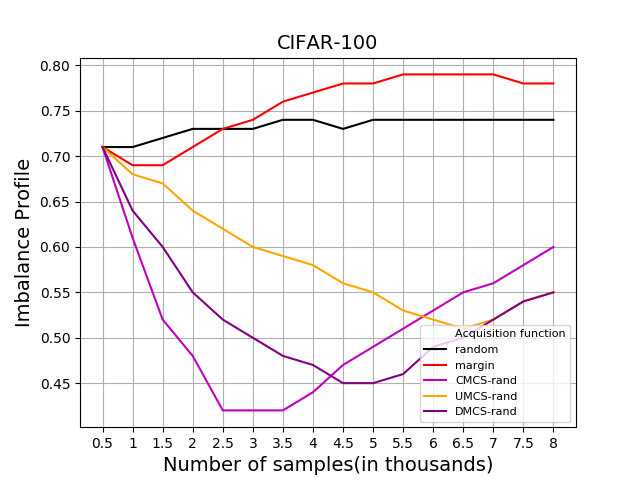}
\includegraphics[width=0.33\textwidth]{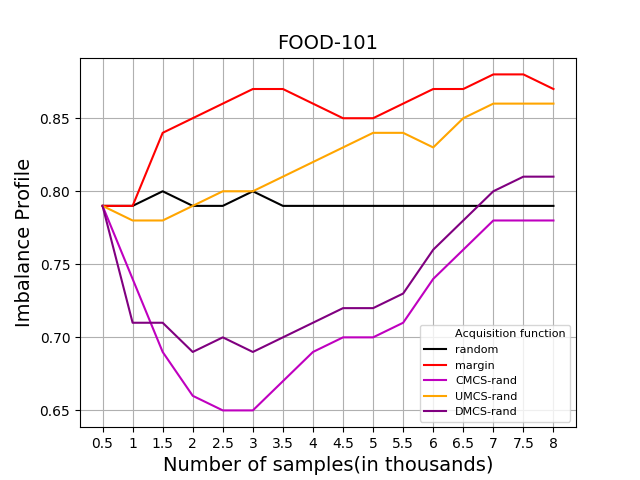}
\includegraphics[width=0.33\textwidth]{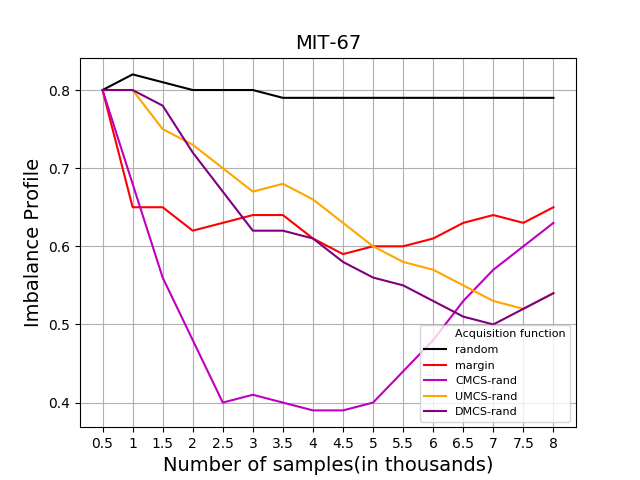}

\vspace{-1em}
    \caption{Imbalance profile of labeled datasets for the three versions of the proposed method. Random sampling is used as AF for remaining samples of each iteration if there are not enough samples associated to minority classes. $b=8000$, $t=16$. \textit{Best viewed in color}.}
    \label{fig:imp_div}
\vspace{-3mm}
\end{figure*}

\subsection{Analysis of minority oriented sampling versions}
The performance and imbalance profiles for three versions of the proposed method described in Subsection~\ref{subsec:method} are presented in Figures~\ref{fig:acc_div} and~\ref{fig:imp_div} respectively. 
The comparison is done with $random$ as auxiliary AF. 
All the three versions have a positive impact in mitigating the imbalance with $DMCS$ providing slightly better global performance.
This finding indicates that a diversified selection of samples for minority classes is better than favoring the most certain or uncertain ones.
Accuracy is slightly better for $DMCS$ and $CMCS$ for CIFAR-100 in the initial iterations, while $UMCS$ is better later. 
This is partly explained by the imbalance profile, which increases for the first two versions but not for the third after a certain iteration. 
For MIT-67, $UMCS$ is most effective since the average performance is higher and the uncertain samples become the ones of interest.
This is also the case of CIFAR-100, but later in the AL process.  

$CMCS$ is most effective to mitigate the imbalance in the early stages, but leads to most imbalanced dataset by the end of the learning process for CIFAR-100 and MIT-67. 
It is likely that $CMCS$ learns a limited representation of the minority class, since it focuses on most certain samples. This reduces the model's ability to find samples for the class and also explain the observation that $CMCS$ is outperformed by $DMCS$ and $UMCS$ in later iterations.

The results for FOOD-101 are particularly interesting since the three selection processes lead to different imbalance profiles while the accuracy is quite similar.  

   


\begin{figure*}[]

\includegraphics[width=0.33\textwidth]{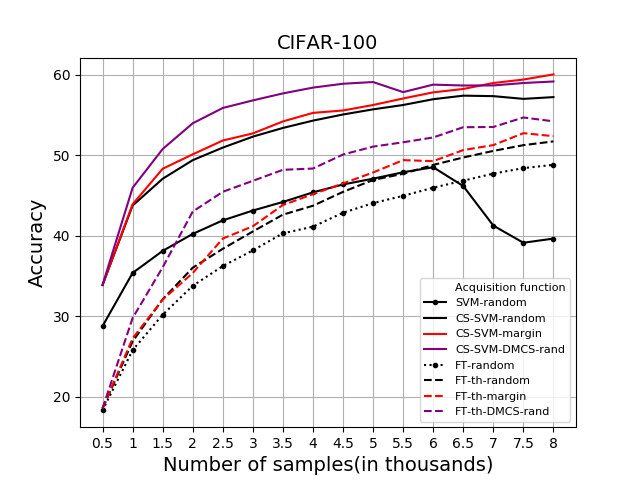}
\includegraphics[width=0.33\textwidth]{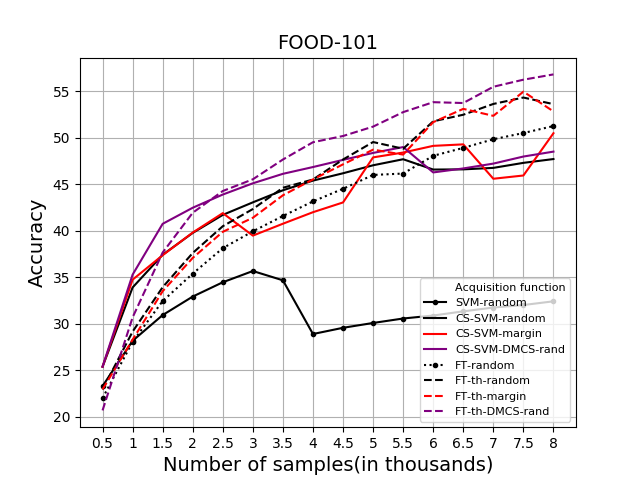}
\includegraphics[width=0.33\textwidth]{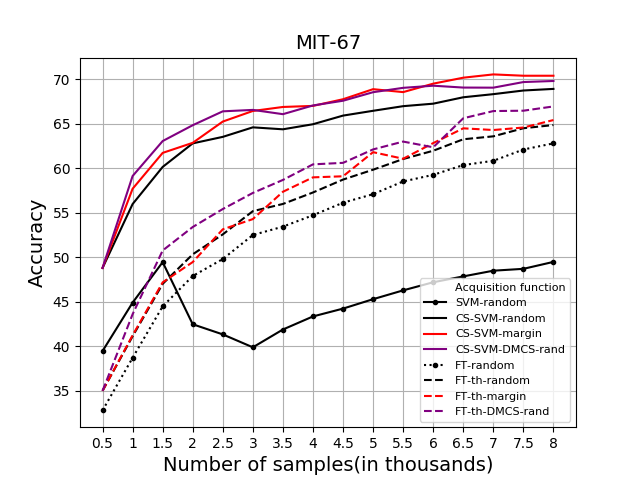}
    
\vspace{-1em}
    \caption{Comparison of classical (SVM, FT) and imbalance-oriented training schemes (CS-SVM, FT, FT-th). $b=8000$, $t=16$.  \textit{Best viewed in color}.}
    \label{fig:ts}
\vspace{-3mm}
\end{figure*}

\subsection{Comparison of training schemes} 
\label{subsec:classical_imbalanced}
We illustrate the results obtained by the two training schemes described in Section~\ref{sec:formalization} without ($SVM$, $FT$) and with ($CS-SVM$, $FT-th$) adaptation for an imbalanced context in Figure~\ref{fig:ts}.
$CS-SVM$ and $FT-th$ outperform $SVM$ and vanilla $FT$, their classical counterparts for $random$. The gain is quite significant for $CS-SVM$, validating its use in class imbalanced active learning~\cite{ertekin2007active}. 
Further, we show the effectiveness of the proposed method $DMCS-rand$ over $random$ and the overall strongest baseline $margin$ for both of the two imbalance adapted schemes($CS-SVM$, $FT-th$). 
This shows the need to explicitly focus on minority classes during the selection process and the effectiveness of the method irrespective of training scheme.
Another interesting remark is that, except for classical $SVM-random$, all other transfer learning based schemes outperform fine-tuning schemes for CIFAR-100 and MIT-67.
Consequently, both training schemes should be tried at the beginning of the AL process for a new unlabeled dataset.
If the distance between a generic dataset and the unlabeled one is not high, transferring features from the first toward the second seems preferable to fine tuning.
Otherwise, fine tuning become better at some point during AL and can replace the transfer learning scheme, as in case of FOOD-101.


\section{Conclusion}
We introduce a new acquisition method which is designed for iterative active learning over imbalanced datasets. 
The method focuses the selection process toward samples which are associated to minority classes in order to reduce the negative effect of imbalance.
Evaluation is performed against competitive baselines and the proposed method ensures a performance gain.
An analysis of its main components facilitates the understanding of their individual contributions.
Surprisingly, we find that transfer learning scheme outperforms the fine tuning based scheme usually deployed in AL. 
We also propose a simple but effective way to test the accuracy of the two schemes after each iteration in order to decide which one should be used later in the AL process.
The results presented here are encouraging and research would be pursued along three axes.
First, the proposed method will be tested on larger datasets to understand its behavior for AL at scale.
Second, the idea to prioritize minority classes can be extended to balanced dataset to favor classes that are difficult to learn.  
Finally the effect of the pretrained dataset on transfer learning will be assessed. 
To do this, ILSVRC will be replaced with a larger dataset, such as the entire ImageNet, for pretraining.

\bibliographystyle{IEEEtran}
\bibliography{ref}

\end{document}